\documentclass[11pt]{article}

\usepackage[]{acl2021}

\usepackage{times}
\usepackage{latexsym}

\usepackage[T1]{fontenc}

\usepackage[utf8]{inputenc}

\usepackage{microtype}
\usepackage{xcolor}
\usepackage{amsmath}
\usepackage{mathtools}
\usepackage{booktabs}
\usepackage{subfigure}
\usepackage{wrapfig}
\usepackage{float}
\restylefloat{table}
\renewcommand{\vec}[1]{\boldsymbol{#1}}
\usepackage[normalem]{ulem}
\usepackage{xspace} 
\usepackage{enumitem}
\usepackage{geometry}

\usepackage{tcolorbox}
\tcbuselibrary{skins,breakable,xparse}
\newcommand{\shading}[1]{%
  \tcbox[tcbox raise base, 
  left=0mm,right=0mm,top=0mm,bottom=0mm,boxsep=0.5pt,arc=0mm,
  boxrule=0pt,opacityfill=0.3,enhanced jigsaw,colback=gray!85!white,
  before=\relax,after=\relax]{\framebox[1.05\width]{\textsc{#1}}}
}

\newcommand{\method}{\textsc{IteraTeR}\xspace}
\usepackage[finalnew]{trackchanges}
\addeditor{VR}
\addeditor{WD}
\addeditor{DK}
\addeditor{ZM}
\addeditor{DH}

\title{Understanding Iterative Revision from Human-Written Text}

\author{Wanyu Du$^1$\thanks{\xspace\xspace This research was performed when Wanyu Du was interning at Grammarly.}\ , Vipul Raheja$^2$\ , Dhruv Kumar$^2$\ , Zae Myung Kim$^3$\ ,\\ {\bf Melissa Lopez$^2$\ ,} {\bf Dongyeop Kang$^4$}\\
$^1$University of Virginia, \space $^2$Grammarly, \\ 
$^3$Univ. Grenoble Alpes, CNRS, LIG, \space $^4$University of Minnesota \\
\texttt{wd5jq@virginia.edu} \\
\texttt{\{vipul.raheja,dhruv.kumar,melissa.lopez\}@grammarly.com} \\
\texttt{zae-myung.kim@univ-grenoble-alpes.fr} \\
\texttt{dongyeop@umn.edu}
}

\date{}

\begin{document}
\maketitle

\begin{abstract}

Writing is, by nature, a strategic, adaptive, and more importantly, an iterative process.
A crucial part of writing is editing and revising the text.
Previous works on text revision have focused on defining edit intention taxonomies within a single domain or developing computational models with a single level of edit granularity, such as sentence-level edits, which differ from human's revision cycles.
This work describes \method: the first large-scale, multi-domain, edit-intention annotated corpus of iteratively revised text.
In particular, \method is collected based on a new framework to comprehensively model the iterative text revisions that generalize to \change[DK]{a variety of domains}{various domains of formal writing}, edit intentions, revision depths, and granularities. 
When we incorporate our annotated edit intentions, both generative and edit-based text revision models significantly improve automatic evaluations.\footnote{Code and dataset are available at \url{https://github.com/vipulraheja/IteraTeR}.}
Through our work, we better understand the text revision process, making vital connections between edit intentions and writing quality, enabling the creation of diverse corpora to support computational modeling of iterative text revisions.
\end{abstract}

\section{Introduction}

Writing is a complex and effortful cognitive task, where writers balance and orchestrate three distinct cognitive processes: planning, translation, and revising \cite{flowerhayes1980}. These processes can be hierarchical and recursive and can occur at any moment during writing. 
This work focuses on text revision as an essential part of writing \cite{scardamalia1986}.
Revising text is a strategic, and adaptive process. It enables writers to deliberate over and organize their thoughts, find a better line of argument, learn afresh, and discover what was not known before \cite{sommers1980revision}. 
Specifically, text revision involves identifying discrepancies between intended and instantiated text, deciding what edits to make, and how to make those desired edits
\cite{faigley1981analyzing, fitzgerald1987research,  bridwell1980}.

\begin{table}[t]
  \centering
  \small
  \begin{tabular}{@{}p{0.48\textwidth}@{}}
    \toprule
    Each comment was annotated by three different annotators, which achieved high inter-annotator agreement. The proposed annotation \{\textcolor{red}{\sout{process}} \textcolor{teal}{approach}\}\shading{Clarity} is also language and domain independent\{\textcolor{teal}{, nevertheless, it was currently applied for Brazilian Portuguese}\}\shading{Meaning-changed}. \\
    \midrule
    Each comment was annotated by three different annotators, \{\textcolor{red}{\sout{which}} \textcolor{teal}{and}\}\shading{Coherence} achieved high inter-annotator agreement. The \{\textcolor{teal}{new}\}\shading{Meaning-changed} proposed annotation approach is also language and \{\textcolor{red}{\sout{domain independent, nevertheless, it was currently}}\textcolor{teal}{domain-independent (although it has been}\} \shading{Clarity} applied for Brazilian Portuguese\{\textcolor{teal}{)}\}\shading{Fluency}. \\
    \midrule
    Each comment was annotated by three different annotators \{\textcolor{red}{\sout{,}}\}\shading{Fluency} and achieved high inter-annotator agreement. The \{\textcolor{red}{\sout{new}}\}\shading{Coherence} proposed annotation approach is also language and domain-independent \{\textcolor{red}{\sout{(although it has been applied}} \textcolor{teal}{nevertheless it is currently customized}\} \shading{Coherence} for Brazilian Portuguese \{\textcolor{red}{\sout{)}}\} \shading{Fluency}.\\
    \bottomrule
  \end{tabular}
  \caption{\label{tab:example}
  An iteratively revised ArXiv abstract snippet (2103.14972, version 2, 3, and 4) with our annotated \shading{edit-intention} in \method.
  \vspace{-4mm}
  }
\end{table}

Text revision is an \textit{iterative} process. 
Human writers are unable to simultaneously comprehend multiple demands and constraints of the task when producing well-written texts \citep{flower1980dynamics, collins1980framework, vaughan-mcdonald-1986-model} -- 
for instance, expressing ideas, covering the content, following linguistic norms and discourse conventions of written prose, etc. 
Thus, they turn towards making \textit{successive iterations of revisions} to reduce the number of considerations at each time. 

Previous works on iterative text revision have three major limitations: 
(1) simplifying the task to \change[DK]{a single-pass}{an noniterative} "original-to-final" text paraphrasing;
(2) focusing largely on sentence-level editing \citep{faruqui-etal-2018-wikiatomicedits,botha-etal-2018-learning,ito-etal-2019-diamonds,faltings-etal-2021-text};
(3) developing editing taxonomies within \change[VR]{singular}{individual} domains (e.g. Wikipedia articles, academic writings) \citep{yang-etal-2017-identifying-semantic,zhang-etal-2017-corpus,anthonio-etal-2020-wikihowtoimprove}. 
These limitations make their proposed text editing taxonomies, datasets, and models\remove[WD]{, and} lose their generalizability and practicality. 

We present \method --- an annotated dataset for \textsc{Itera}tive \textsc{Te}xt \textsc{R}evision that consists of 31,631 iterative document revisions with sentence-level and paragraph-level edits across multiple domains \add[DK]{of formally human-written text}, including Wikipedia\footnote{\url{https://www.wikipedia.org/}}, ArXiv\footnote{\url{https://arxiv.org/}} and Wikinews.\footnote{\url{https://www.wikinews.org/}}
Table \ref{tab:example} shows a sample ArXiv document in \method, that underwent iterative revisions.
Our dataset includes 4K \change[WD]{human-}{manually }annotated and 196K automatically\change[WD]{-}{ }annotated edit intentions based on a sound taxonomy we developed, and is generally applicable across multiple domains and granularities (See Table \ref{tab:datasets}).
Note that \textsc{IteraTeR} is currently only intended to support formal writing revisions, as iterative revisions are more prevalent in formal rather than informal  writings (e.g. tweets, chit-chats)\footnote{Further extension to less formal writings (e.g. blog, emails) will be discussed in the future.}.
Our contributions are as follows:

\begin{itemize}[noitemsep,topsep=0pt,leftmargin=*]
    \item formulate the iterative text revision task in a more comprehensive way, capturing greater real-world challenges such as successive revisions, multi-granularity edits, and domain shifts. 
    \item collect and release a large, multi-domain Iterative Text Revision dataset: \method, which contains 31K document revisions from Wikipedia, ArXiv and Wikinews, and 4K edit actions with high-quality edit intention annotations.
    \item analyze how text quality evolves across iterations and how it is affected by different kinds of edits.
    \item \change[VR]{find}{show} that incorporating the annotated edit-intentions is advantageous for text revision systems to generate better-revised documents.
\end{itemize}
\section{Related Work}

\paragraph{Edit Intention Identification.}
Identification of edit intentions is an integral part of the iterative text revision task. Prior works have studied the categorization of different types of edit actions to help understand why editors do what they do and how effective their actions are \citep{yang-etal-2017-identifying-semantic,zhang-etal-2017-corpus,ito-etal-2019-diamonds}. 
However, these works do not further explore how to leverage edit intentions to generate better-revised documents. 
Moreover, some of their proposed edit intention taxonomies are constructed with a focus on specific domains of writing, such as Wikipedia articles \citep{anthonio-etal-2020-wikihowtoimprove,bhat-etal-2020-towards,faltings-etal-2021-text} or academic essays \citep{zhang-etal-2017-corpus}. As a result, their ability to generalize to other domains remains an open question.


\begin{table}[t]
  \centering
  \small
  \begin{tabular}{@{}p{0.20\textwidth}@{\hskip 2mm}l@{\hskip 1mm}c@{\hskip 1mm}c@{\hskip 1mm}c@{\hskip 1mm}c@{}}
    \toprule
    \textbf{Dataset} & \textbf{Size} & \textbf{Domain} & \textbf{Gran.} & \textbf{Hist.} & \textbf{Ann.} \\
    \midrule
    \citet{yang-etal-2017-identifying-semantic} & 5K & Wiki & P & $\times$ & $\surd$ \\
    \citet{anthonio-etal-2020-wikihowtoimprove} & 2.7M & Wiki & S & $\surd$ & $\times$ \\
    \citet{zhang-etal-2017-corpus} & 180 & Academic & S & $\surd$ & $\surd$  \\
    \citet{DBLP:journals/corr/abs-2104-09647} & 4.6M & News & S & $\surd$ & $\times$   \\
    \method (Ours) & 31K & All & S\&P & $\surd$ & $\surd$  \\
    \bottomrule
  \end{tabular}
  \caption{\label{tab:datasets}
  Comparisons with previous related works. Gran. for Granularity: S for sentence-level and P for paragraph-level. Hist. for Revision History. Ann. for Edit Intention Annotations. 
  \vspace{-4mm}
  }
\end{table}

\paragraph{\change[DK]{Single-Pass}{Noniterative} Text Revision Models.}
Some prior works \citep{faruqui-etal-2018-wikiatomicedits,botha-etal-2018-learning,ito-etal-2019-diamonds,faltings-etal-2021-text}
simplify the text revision task to a single-pass "original-to-final" sentence-to-sentence generation task. 
However, it is very challenging to conduct multiple perfect edits at once. 
For example, adding transition words or reordering the sentences are required to further improve the document quality. 
Therefore, single-pass sentence-to-sentence text revision models are not sufficient to deal with real-world challenges of text revision tasks.
In this work, we explore the performance of text revision models in multiple iterations and multiple granularities.

\paragraph{Iterative Text Revision Datasets.}
While some prior works have constructed iterative text revision datasets, they are limited to singular writing domains, such as Wikipedia-style articles \citep{anthonio-etal-2020-wikihowtoimprove}, academic essays \citep{zhang-etal-2017-corpus} or news articles \citep{DBLP:journals/corr/abs-2104-09647}.
In this work, we develop a unified taxonomy to analyze the characteristics of iterative text revision behaviors across different domains and collect large scale text revisions of human writings from multiple domains.
The differences between \method and the prior datasets are summarized in Table \ref{tab:datasets}.

\begin{table*}[ht]
  \centering
  \small
  \begin{tabular}{crr|rr|rr|rr|rr|rr}
    \toprule
    \multicolumn{1}{l}{} & \multicolumn{6}{c|}{\textbf{\textsc{IteraTeR-full}}} & \multicolumn{6}{c}{\textbf{\textsc{IteraTeR-human}}}\\
    \cmidrule(lr){2-7} \cmidrule(lr){8-13}
    \multicolumn{1}{l}{} & \multicolumn{2}{c|}{\textbf{ArXiv}} & \multicolumn{2}{c|}{\textbf{Wikipedia}} & \multicolumn{2}{c|}{\textbf{Wikinews}} & \multicolumn{2}{c|}{\textbf{ArXiv}} & \multicolumn{2}{c|}{\textbf{Wikipedia}} & \multicolumn{2}{c}{\textbf{Wikinews}}  \\
    \cmidrule(lr){2-3} \cmidrule(lr){4-5} \cmidrule(lr){6-7} \cmidrule(lr){8-9} \cmidrule(lr){10-11} \cmidrule(lr){12-13}
    \textbf{Depth} & \#D & \#E & \#D & \#E & \#D & \#E  & \#D & \#E  & \#D & \#E  & \#D & \#E \\
    \midrule
    1 & 9,446 & 65,450 & 8,195 & 51,290 & 7,878 & 39891 & 95 & 618 & 130 & 1,072 & 173 & 1,227 \\
    2 & 1,615 & 11,391 & 1,991 & 12,868 & 1,455 & 8,116 & 76 & 499 & 38 & 250 & 25 & 155 \\
    3 & 301 & 2,076 & 415 & 2,786 & 161 & 1,704 & 6 & 47 & 10 & 98 & 4 & 27 \\
    4 & 66 & 444 & 64 & 723 & 16 & 71 & 1 & 13 & 1 & 12 & 0 & 0 \\
    5 & 15 & 107 & 9 & 52 & 4 & 18 & 0 & 0 & 0 & 0 & 0 & 0 \\
    \midrule
    \textbf{Total} & 11,443 & 79,468 & 10,674 & 67,719 & 9,514 & 49,800 & 178 & 1,177 & 179 & 1,432 & 202 & 1,409 \\
    \bottomrule
  \end{tabular}
  \caption{\label{tab:revision-depth}
  Statistics of the \method dataset, where \#D indicate the number of document revisions ($\mathcal{R}^t$), and \#E indicate the number of annotated edit actions.
  }
\end{table*}

\section{Formulation: Iterative Text Revision}

We provide formal definitions of the Iterative Text Revision task, and its building blocks.

\paragraph{Edit Action.}
An edit action $\vec{a}_k$ is a local change applied \change[WD]{on}{to} a certain text object, where $k$ is the index of the current edit action.
The local changes include: insert, delete and modify.
The text objects include: token, phrase\add[WD]{\footnote{In this work, we define phrase as text pieces which contain more than one token and only appears within a sentence.}}, sentence, and paragraph.
This work defines local changes applied \change[WD]{on}{to} tokens or phrases as \textit{sentence-level edits}, local changes applied \change[WD]{on}{to} sentences as \textit{paragraph-level edits} and local changes applied \change[WD]{on}{to} paragraphs as \textit{document-level edits}.

\paragraph{Edit Intention.}
An edit intention $\vec{e}_k$ reflects the revising goal of the editor when making a certain edit action.
In this work, we assume each edit action $\vec{a}_k$ will only be labeled with one edit intention $\vec{e}_k$.
We further describe our edit intention taxonomy in Table \ref{tab:intents-satistics} and \S\ref{sec:dataset_taxonomy}.

\paragraph{Document Revision.}
A document revision is created when an editor saves changes for the current document  \citep{yang2016did,yang-etal-2017-identifying-semantic}. 
One revision $\mathcal{R}^t$ is aligned with a pair of documents $(\mathcal{D}^{t-1}, \mathcal{D}^{t})$ and contains \change[WD]{$K$}{$K^t$} edit actions, where $t$ indicates the version of the document and \change[WD]{$K\ge1$}{$K^t \ge1$}.
A revision with \change[WD]{$K$}{$K^t$} edit actions will correspondingly have \change[WD]{$K$}{$K^t$} edit intentions:
\begin{equation}
    (\mathcal{D}^{t-1}, \mathcal{D}^{t}) \to \mathcal{R}^t=\{(\vec{a}^t_k, \vec{e}^t_k)\}_{k=1}^{K^t}
\end{equation}
We define $t$ as the \remove[WD]{document version or }revision depth\remove[WD]{, which we use interchangeably}.

\paragraph{Iterative Text Revision.}
Given a source text $\mathcal{D}^{t-1}$, iterative text revision is the task of generating revisions of text $\mathcal{D}^{t}$ at depth $t$ until the quality of the text in the final revision satisfies a set of pre-defined stopping criteria $\{s_0, ..., s_M\}$:
\begin{equation}
     \mathcal{D}^{t-1} \xrightarrow[]{g(\mathcal{D})} \mathcal{D}^{t}, \text{if} ~ f(\mathcal{D}^{t}) < \{s_0, ..., s_M\}
\end{equation}
where $g(\mathcal{D})$ is a text revision system and $f(\mathcal{D})$ is a quality evaluator of the revised text.
The quality evaluator $f(\mathcal{D})$ can be automatic systems or manual judgements which measure the quality of the revised text. 
\add[WD]{The stop criteria $\{s_i\}$ is a set of conditions that determine whether to continue revising or not.}
In this work, we simply set them as revision depth equal to 10, and edit distance between $\mathcal{D}^{t-1}$ and $\mathcal{D}^t$ equal to 0 (\S\ref{sec:iterativeness-exp}).
\add[WD]{We will include other criteria which measures the overall quality, content preservation, fluency, coherence and readability of the revised text in future works.}

\begin{table*}[t]
  \centering
  \small
  \begin{tabular}{@{}r@{\hskip 1mm}p{0.38\textwidth}@{\hskip 1mm}p{0.27\textwidth}@{\hskip 1mm}r@{\hskip 1mm}}
    \toprule
     \textbf{Edit-Intention} & \textbf{Description} & \textbf{Example} & \textbf{Counts (Ratio)} \\
    \toprule
    \shading{Fluency} & Fix grammatical errors in the text. & She went to the \textcolor{red}{\sout{markt}}\textcolor{teal}{market}. & 942 (23.44\%) \\
    \midrule
    \shading{Coherence} & \change[WD]{Making}{Make} the text more cohesive, logically linked and consistent as a whole. & She works hard\textcolor{red}{\sout{. She}}\textcolor{teal}{; therefore, she} is successful. & 393 (9.78\%) \\
    \midrule
    \shading{Clarity} & \change[WD]{Edits to make}{Make} the text more \change[WD]{clear}{formal, concise}, readable and understandable. & The changes \textcolor{red}{\sout{made the paper better than before}}\textcolor{teal}{improved the paper}. & 1,601 (39.85\%) \\
    \midrule
    \shading{Style} & \change[WD]{Edits to justly convey}{Convey} the writer’s \remove[WD]{context, emotions, tone, formality, voice, and other} writing preferences, \add[WD]{including emotions, tone, voice, etc.}. & Everything was \textcolor{teal}{awfully} rotten. & 128 (3.19\%) \\
    \midrule
    \shading{Meaning-Changed} & Update or add new information to the text. & This method improves the model accuracy from 64\% to \textcolor{red}{\sout{78}}\textcolor{teal}{83}\%. & 896 (22.30\%) \\
    \midrule
    \midrule
    \shading{Other} & Edits that are not recognizable and do not belong to the above intentions. & This method is also named as \textcolor{red}{\sout{CITATION1}}. & 58 (1.44\%) \\
    \bottomrule
  \end{tabular}
  \caption{\label{tab:intents-satistics}
  A taxonomy of edit intentions in \method, where \shading{Fluency}, \shading{Coherence}, \shading{Clarity} and \shading{Style} belong to \shading{Non-Meaning-changed} edits.
  \vspace{-3mm}
  }
\end{table*}

\section{\method Dataset}
\label{sec:dataset}

\subsection{Raw Data Collection}
\label{sec:dataset_collection}

\paragraph{Domains.}
We select three domains -- Wikipedia articles, academic papers, and news articles -- to cover different human writing goals, formats, revision patterns, and quality standards. 
The three domains consist of formally written texts, typically edited by multiple authors. 
We describe \add[WD]{why and }how we collect \change[WD]{revision text}{text revision} from each domain below:
\begin{itemize}[noitemsep,topsep=0pt,leftmargin=*]
\item \textbf{Scientific Papers.}
Scientific articles are written in a rigorous, logical manner. 
Authors generally highlight and revise their hypotheses, experimental results, and research insights in this domain. 
We collect paper abstracts submitted at different timestamps (i.e., version labels) from ArXiv. 
\item \textbf{Wikipedia Articles.}
Encyclopedic articles are written in a formal, coherent manner, where editors typically focus on improving the clarity and structure of articles to make people easily understand all kinds of factual and abstract encyclopedic information. 
We collect revision histories of the main contents of Wikipedia articles.
\item \textbf{News Articles.}
News articles are generally written in a precise and condensed way. News editors emphasize improving the clarity and readability of news articles to keep people updated on rapidly changing news events. We collect revision histories of news content from Wikinews.
\end{itemize}

\paragraph{Raw Data Processing.}
We first collect all raw documents, then sort each document version according to its timestamp in ascending order.
For each document $\mathcal{D}$, we pair two consecutive versions as one revision $(\mathcal{D}^{t-1}, \mathcal{D}^{t})\to \mathcal{R}^t$, where $t$ is the revision depth.
For each sampled document-revision $\mathcal{R}^t$, we extract its full edit actions using \textit{latexdiff}.\footnote{\url{https://www.ctan.org/pkg/latexdiff}} 
We provide both the paragraph-level and sentence-level revisions where the latter is constructed by applying a sentence segmentation tool,\footnote{\url{https://github.com/zaemyung/sentsplit}} and aligning each sentence to each revision.
For each revision pair, we have: the revision type, the document id, the revision depth, an original phrase and a revised phrase, respectively.\footnote{We also record character-level indices of their positions within the original sentence and the paragraph.}
The detailed processing of raw text is described in Appendix \ref{appendix:data_processing}.

In summary, we collect 31,631 document revisions with 196,987 edit actions, and maintain a relatively balanced distribution across three domains, as shown in Table \ref{tab:revision-depth}. We call this large-scale dataset as \textsc{IteraTeR-full-raw}.

\subsection{Data Annotation}
\label{sec:data_annotation}

To better understand the human revision process, we sample 559 document revisions from \textsc{IteraTeR-full-raw}, consisting of 4,018 human edit actions. We refer to this small-scale unannotated dataset as \textsc{IteraTeR-human-raw}. 
In \S\ref{sec:human_annotation}, we then use Amazon Mechanical Turk (AMT) to crowdsource edit intention annotations for each edit action according to our proposed edit-intention taxonomy (\S\ref{sec:dataset_taxonomy}). 
We refer to this small-scale annotated dataset as \textsc{IteraTeR-human}.\footnote{We provide our annotation instruction in Appendix \ref{sec:annotation-interface}.}

We then scale these \change[WD]{human}{manual} annotations to \textsc{IteraTeR-full-raw} by training edit intention prediction models on \textsc{IteraTeR-human}, and automatically label \textsc{IteraTeR-full-raw} to construct \textsc{IteraTeR-full}. (\S\ref{sec:auto_annotation})

\subsubsection{Edit Intention Taxonomy}
\label{sec:dataset_taxonomy}
For manual annotations, we propose a new edit intention taxonomy in \method (Table \ref{tab:intents-satistics}), in order to comprehensively model the iterative text revision process. 
Our taxonomy builds on prior literature \citep{6448848, harris2017writing}. 
At the highest level, we categorize the edit intentions into ones that change the meaning or the information contained in the text (\textsc{Meaning-changed}), and ones that preserve these characteristics (\textsc{Non-Meaning-changed}). 
Since our goal is to understand edit intentions to improve the quality of writing, we focus on categorizing edits in the latter category further into four sub-categories: \textsc{Fluency}, \textsc{Clarity}, \textsc{Coherence} and \textsc{Style}. Our proposed taxonomy of edit intentions is generally applicable to multiple domains, edit-action granularities (sentence-level and paragraph-level), and revision depths. We also propose the \textsc{Other} category for edits that cannot be labeled using the above taxonomy.


\begin{figure*}[t]
    \centering
    \includegraphics[width=1.02\textwidth]{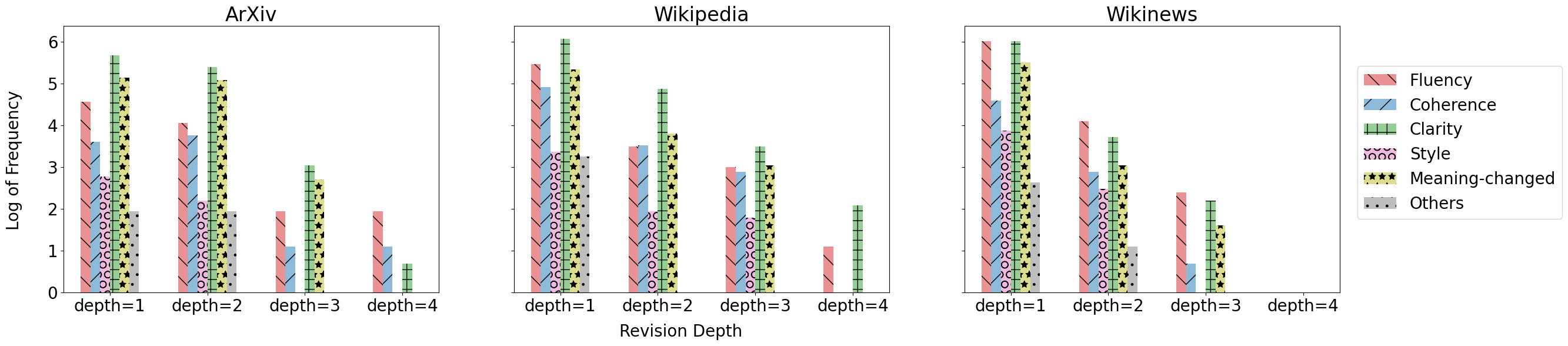}
    \caption{Logarithm (base $e$) of frequency for edit-intentions in each revision depth for the three dataset domains.}
    \label{fig:edits_per_domain}
\end{figure*}

\subsubsection{Manual Annotation}
\label{sec:human_annotation}

Since edit intention annotation is a challenging task, we design strict qualification tests to select 11 qualified AMT annotators (details in Appendix \ref{appendix:qualification_tests}).
To further improve the annotation quality, we ask another group of expert linguists (English L1, bachelor's or higher degree in Linguistics) to re-annotate the edits which do not have a majority vote among the AMT workers.
Finally, we take the majority vote among 3 human annotations (either from AMT workers or from expert linguists) as the final edit intention labels.
This represents the \textsc{IteraTeR-human} dataset. We release both the final majority vote and the three raw human annotations per edit action as part of the dataset.

\subsubsection{Automatic Annotation}
\label{sec:auto_annotation}
To scale up the annotation, we train an edit-intention classifier to annotate \textsc{IteraTeR-full-raw} and construct the \textsc{IteraTeR-full} dataset. 
We split the \textsc{IteraTeR-human} dataset into 3,254/400/364 \change[WD]{train, validation and testing}{training, validation and test} pairs.
The edit intention classifier is a RoBERTa-based \cite{liu2019roberta} multi-class classifier that predicts an intent given the original and the revised text for each edit action\footnote{Please refer to Appendix \ref{appendix:auto_ann_details} for more training details.}.
Table \ref{tab:intent-classifier} shows its performance on the test set. The Fluency and Clarity edit intentions are easy to predict with F1 scores of 0.8 and 0.69, respectively, while Style and Coherence edit intentions are harder to predict with F1 scores of 0.13 and 0.32, respectively, largely due to the limited occurrence of Style and Coherence intents in the training data (Table \ref{tab:intents-satistics}).

\subsection{Data Analysis}
\begin{table}[t]
  \centering
  \small
  \begin{tabular}{@{}lcccc@{}}
    \toprule
    \textbf{Edit-Intention} & \textbf{Precision} & \textbf{Recall} & \textbf{F1} \\
    \midrule 
    \shading{Clarity} & 0.75 & 0.63 & 0.69  \\
    \shading{Fluency} & 0.74 & 0.86 & 0.80  \\
    \shading{Coherence} & 0.29 & 0.36 & 0.32 \\
    \shading{Style} & 1.00 & 0.07 & 0.13 \\
    \shading{Meaning-Changed} & 0.44 & 0.69 & 0.53  \\
    \bottomrule
  \end{tabular}
  \caption{
  \label{tab:intent-classifier}
  Edit intention classifier performance on the test split of \textsc{IteraTER-human}.
  \vspace{-4mm}
  }
\end{table}

\paragraph{Edit Intention Distributions.}
The iterative edit intention distributions \change[WD]{under}{in} three domains are demonstrated in Figure \ref{fig:edits_per_domain}.
Across all three domains, authors tend to make the majority of edits at revision depth 1. However, the number of edits rapidly decreases at revision depth 2, and few edits are made at revision depth 3 and 4.

We find that \textsc{Clarity} is one of the most frequent edit intention\add[WD]{s} across all domains, indicating that authors focus on improving readability across all domains.
For ArXiv, \textsc{Meaning-changed} edits are also among the most frequent edits, which indicates that authors also focus on updating the contents of their abstracts to share new research insights or update existing ones.
\add[WD]{Meanwhile, ArXiv also covers many \textsc{Fluency} and \textsc{Coherence} edits, collecting edits from scientific papers and suggesting meaningful revisions would be an important future application of our dataset. }
For Wikipedia, we find that \textsc{Fluency}, \textsc{Coherence}, and \textsc{Meaning-changed} edits roughly share a similar frequency, which indicates Wikipedia articles have more complex revision patterns than ArXiv and news articles.
For Wikinews, \textsc{Fluency} edits are equally emphasized, indicating that improving grammatical correctness of the news articles is just as important.

\begin{table}[t]
  \centering
  \small
  \begin{tabular}{ccccc}
    \toprule
    & \textbf{ArXiv} & \textbf{Wikipedia} & \textbf{Wikinews} & \textbf{All} \\
    \midrule
    1st-round & 0.3369 & 0.3630 & 0.3886 & 0.3628 \\
    2nd-round & 0.4983 & 0.4274 & 0.5601 & 0.5014 \\
    \bottomrule
  \end{tabular}
  \caption{\label{tab:iaa}
  Inter-annotator agreement (Fleiss' $\kappa$ \citep{fleiss1971measuring}) across two rounds of annotations, where the 1st-round only contains annotations from qualified AMT workers, and the 2nd-round contains annotations from both qualified AMT workers and expert linguists.
  \vspace{-3mm}
  }
\end{table}

\paragraph{Inter-Annotator Agreement.}
We measure inter-annotator agreement (IAA) using the Fleiss' $\kappa$ \citep{fleiss1971measuring}. Table \ref{tab:iaa} shows the IAA across three domains.
After the second round of re-annotation by proficient linguists, the Fleiss' $\kappa$ increases to 0.5014, which indicates moderate agreement among annotators.

We further look at the raw annotations where at least 1 out of 3 annotators assign\add[WD]{s} a different edit intention label. 
We find that the \textsc{Coherence} intention is the one that is the most likely to have a disagreement: 312 out of 393 \textsc{Coherence} annotations do not have consensus. 
Within those disagreements of the \textsc{Coherence} intention, 68.77\% are considered to be \textsc{Clarity}, and 11.96\% are considered to be the \textsc{Fluency} intention.
Annotators also often disagree on the \textsc{Clarity} intention, where 1023 out of 1601 \textsc{Clarity} intentions do not have a consensus. 
Among those disagreements of the \textsc{Clarity} intention, 30.33\% are considered to be \textsc{Coherence}, and 30.23\% are considered to be \textsc{Style}.

The above findings explain why the inter-annotator agreement scores are lower in Wikipedia and ArXiv. 
As shown in Figure \ref{fig:edits_per_domain}, Wikipedia has many \textsc{Coherence} edits while ArXiv has many \textsc{Clarity} edits.
This explains the difficulty of the edit intention annotation task: it not only asks annotators to infer the edit intention from the full document context, but also requires annotators to have a wide range of domain-specific knowledge in scientific writings.

\section{Understanding Iterative Text Revisions}\label{sec:understanding_human}

To better understand how \change[WD]{human}{text} revisions affect the overall quality of documents, we conduct both \change[WD]{human}{manual} and automatic evaluations on a sampled set of document revisions.

\subsection{Experiment Setups}
\label{sec:human-data-setup}
\paragraph{Evaluation Data.}
We sample two sets of text revisions for different evaluation purposes.
The first set contains 21 iterative document revisions, consisting of 7 unique documents, each document having 3 document revisions from revision depth 1 to 3. 
The second set contains 120 text pairs, each associated with exactly one edit intention of \textsc{Fluency}, \textsc{Coherence}, \textsc{Clarity} or \textsc{Style}.
We validate the following research questions:
\begin{enumerate}[noitemsep,topsep=0pt,leftmargin=*,label=RQ\arabic*]
    \item How do human revisions affect the text quality across revision depths?
    \item How does text quality vary across edit intentions?
\end{enumerate}

\paragraph{Manual Evaluation Configuration.}
We hire a group of proficient linguists to evaluate the overall quality of the documents/sentences, where each revision is annotated by 3 linguists.
For each revision, we randomly shuffle the original and \remove[WD]{human }revised texts, and ask the evaluators to select which one has better overall quality. They can choose one of the two texts, or neither.
Then, we calculate the score for the overall quality of the human revisions as follows: 
-1 means the \change[WD]{human revision}{revised text} has worse overall quality than the original text; 
0 means the \change[WD]{human revision}{revised text} do not show a better overall quality than the original text, or cannot reach agreement among 3 annotators;
1 means the \change[WD]{human revision}{revised text} has better overall quality than the original text.

\paragraph{Automatic Evaluation Configuration.}
We select four automatic metrics to \change[WD]{evaluate}{measure} the document quality on four different aspects:  Syntactic Log-Odds Ratio (SLOR) \citep{kann-etal-2018-sentence} for \change[WD]{Fluency}{text fluency evaluation}, Entity Grid (EG) score \citep{lapata2005automatic} for \change[WD]{Coherence}{text coherence evaluation}, Flesch–Kincaid Grade Level (FKGL) \citep{kincaid1975derivation} for \change[WD]{Readability}{text readability evaluation} and BLEURT score \citep{sellam-etal-2020-bleurt} for \change[WD]{Content Preservation}{content preservation evaluation}.
We describe the detailed justification of our metric selection in Appendix \ref{appendix:auto_eval_metrics_justification}.
\add[WD]{However, in our following experiments, we find these existing automatic metrics are poorly correlate with manual evaluations. }

\subsection{Quality Analyses on Revised Texts}

\paragraph{RQ1: Iterative Revisions vs. Quality.}
Table \ref{tab:iterative_revision_metrics} shows the document quality changes at different revision depths.
Generally, human revisions improve the overall quality of original documents, as indicated by the overall score at each revision depth.\footnote{We further validate this observation in another set of 50 single document-revisions in Appendix \ref{sec:human_eval_single_doc}.}
However, the overall quality keeps decreasing as the revision depth increases from 1 to 3, likely because it is more difficult for evaluators to grasp the overall quality in the deeper revision depths in the pair-wise comparisons between the original and revised documents\add[WD]{, because less \textsc{Non-Meaning-changed} edits have been conducted in deeper revision depths}.
For automatic metrics, 
we find $\Delta$SLOR and $\Delta$EG are not well-aligned with human overall score, we further examine whether human revisions makes original documents less fluent and less coherent in the analysis of RQ2.

\begin{table}[t]
  \centering
  \scriptsize	
  \begin{tabular}{@{}c|c|cccc@{}}
    \toprule
    \textbf{$t$} & \textbf{Overall} $\uparrow$ & \textbf{BLEURT}$\uparrow$ & \textbf{$\Delta$SLOR} $\uparrow$ & \textbf{$\Delta$EG} $\uparrow$ & \textbf{$\Delta$FKGL} $\downarrow$\\
    \midrule
    1 & 0.4285 & 0.1982 & -0.0985 & -0.0132 & -1.0718 \\
    2 & 0.4285 & 0.1368 & -0.1025 & -0.0295 & -2.4973 \\
    3 & 0.1428 & -0.0224 & -0.0792 & 0.0278 & 1.8131 \\
    \bottomrule
  \end{tabular}
  \caption{\label{tab:iterative_revision_metrics}
  Evaluation results for 21 iterative document revisions, where $t$ indicates the revision depth.
  Note that $\Delta$SLOR, $\Delta$EG and $\Delta$FKGL are computed by subtracting the scores of original documents from the scores of revised documents. 
  Overall is the manual evaluation of overall quality of the revised documents.
  \vspace{-3mm}
  }
\end{table}

\begin{table}[t]
  \centering
  \small
  \begin{tabular}{@{}cccc@{}}
    \toprule
    \shading{Fluency} & \shading{Coherence} & \shading{Clarity} & \shading{Style} \\
    \midrule
    0.3673 & 0.1500 & 0.2800 & -0.0385 \\
    \bottomrule
  \end{tabular}
  \caption{\label{tab:single_sents_quality}
  Manually evaluated text quality of 120 single sentence-level edits for different edit intentions.\vspace{-3mm}
  }
\end{table}
\paragraph{RQ2: Edit Intentions vs. Quality.}
Table \ref{tab:single_sents_quality} shows how text quality varies across edit intentions.
We find that \textsc{Fluency} and \textsc{Coherence} edits indeed improve the overall quality of original sentences according to human judgments.
This finding suggests that $\Delta$SLOR and $\Delta$EG are not well-aligned with human judgements, and calls for the need to explore other effective automatic metrics to evaluate \change[WD]{Fluency and Coherence in documents}{the fluency and coherence of revised texts}.
Besides, we observe that \textsc{Style} edits degrade the overall quality of original sentences.
This observation also makes sense since \textsc{Style} edits reflect the writer's personal \change[WD]{manner}{writing preferences} (according to our edit intention taxonomy in Table \ref{tab:intents-satistics}), which not necessarily improve the readability, fluency or coherence of the text. 
\section{Modeling Iterative Text Revisions}
\label{sec:models}
To better understand the challenges of modeling the task of iterative text revisions, we train different types of text revision models using \method. 

\subsection{Experiment Setups}

\paragraph{Text Revision Models.}
For training the text revision models, we experiment with both edit-based and generative models. For the edit-based model, we use \textsc{Felix} \cite{mallinson-etal-2020-felix}, and for the generative models, we use \textsc{BART} \cite{lewis-etal-2020-bart} and \textsc{Pegasus} \cite{zhang2020pegasus}. \textsc{Felix} decomposes text revision into two sub-tasks: Tagging, which uses a pointer mechanism to select the subset of input tokens and their order; and Insertion, which uses a masked language model to \change[WD]{in-fill}{fill in} missing tokens in the output not present in the input. \change[ZM]{\textsc{Pegasus}, on the other hand, is a large transformer-based encoder-decoder model pre-trained for abstractive summarization.}{\textsc{BART} and \textsc{Pegasus} are Transformer-based encoder-decoder models which are used in a wide range of downstream tasks such as natural language inference, question answering, and summarization.}

\paragraph{Training.}
We use four training configurations to evaluate whether edit intention information can help better model text revisions. The first configuration uses the pure revision pairs without edit intention annotations (\textsc{IteraTeR-human-raw} dataset). In the second configuration, we include the manually annotated edit intentions to the source text (\textsc{IteraTeR-human} dataset). Similarly, for the third and fourth training configurations, we use \textsc{IteraTeR-full-raw} dataset (no edit intention information) and \textsc{IteraTeR-full} dataset  (automatically annotated labels, as described in \S\ref{sec:auto_annotation}, simply appended to the input text). We use these four configurations for all model architectures.

\begin{table}[t]
  \centering
  \small
  \begin{tabular}{@{}l@{\hskip 2mm}lcccc@{}}
    \toprule
    \textbf{Model} & \textbf{Dataset} & \textbf{SARI} & \textbf{BLEU} & \textbf{R-L} & \textbf{Avg.}  \\
    \midrule
    \textsc{FELIX} & \textsc{human-raw} & 29.23 & 49.48 & 63.43 & 47.38 \\
    \textsc{FELIX} & \textsc{human} & 30.65 & 54.35 & 59.06 & 48.02 \\
    \textsc{FELIX} & \textsc{full-raw} & 30.34 & 55.10 & 56.49 & 47.31 \\
    \textsc{FELIX} & \textsc{full} & \textbf{33.48} & \textbf{61.90} & \textbf{63.72} & \textbf{53.03} \\    
    \midrule
    \midrule
    \textsc{BART} & \textsc{human-raw} & 33.20 & \textbf{78.59} & 85.20 & 65.66 \\
    \textsc{BART} & \textsc{human} & 34.77 & 74.43 & 84.45 & 64.55 \\
    \textsc{BART} & \textsc{full-raw} & 33.88 & 78.55 & 86.05 & 66.16 \\
    \textsc{BART} & \textsc{full} & \textbf{37.28} & 77.50 & \textbf{86.14} & \textbf{66.97} \\
    \midrule
    \textsc{Pegasus} & \textsc{human-raw} & 33.09 & \textbf{79.09} & 86.77 & 66.32 \\
    \textsc{Pegasus} & \textsc{human} & 34.43 & 78.85 & 86.84 & 66.71 \\
    \textsc{Pegasus} & \textsc{full-raw} & 34.67 & 78.21 & \textbf{87.06} & 66.65 \\
    \textsc{Pegasus} & \textsc{full} & \textbf{37.11} & 77.60 & 86.84 & \textbf{67.18} \\
    \midrule
    Baseline & - & 29.47 & 81.25 & 88.04 & 66.25 \\
    \bottomrule
  \end{tabular}
  \caption{\label{tab:model-performances-sari-bleu-rl-avg}
  Model performances on the test set of \textsc{IteraTeR-human}.  Baseline refers to a no-edit baseline, where we simply use the input text as the output. \add[WD]{\textbf{Avg.} is the average score of \textbf{SARI}, \textbf{BLEU} and \textbf{R-L}.}
  }
\end{table}

\subsection{Results Analysis}
\label{sec:iterativeness-exp}
\paragraph{Automatic Evaluation.}
Table \ref{tab:model-performances-sari-bleu-rl-avg} shows the results of the \change[ZM]{two}{three} models for our different training configurations.
Following prior works \citep{malmi-etal-2019-encode,dong-etal-2019-editnts,mallinson-etal-2020-felix}, we report SARI, BLEU, and ROUGE-L metrics, and include detailed breakdown of scores in Appendix \ref{sec:auto_eval_details}. 
It is noteworthy that the SARI score on the no-edit baseline is the lowest, which indicates the positive impact of revisions on document quality, as also corroborated by the human evaluations in \S\ref{sec:understanding_human}.
For both \textsc{IteraTeR-human} and \textsc{IteraTeR-full} datasets, we see that edit intention annotations help to improve the performance of both \textsc{Felix} and \textsc{Pegasus}.
Also, both models perform better on the larger \textsc{IteraTeR-full} dataset compared to the \textsc{IteraTeR-human} dataset, showing that the additional data (and automatically-annotated annotations) are helpful.

\begin{table}[t]
  \centering
  \small 
  \begin{tabular}{@{}lc@{\hskip 2mm}c@{\hskip 2mm}c@{}}
    \toprule
     & \textbf{Human Revision} & \textbf{Tie} & \textbf{Model Revision} \\
    \midrule
    \textbf{Overall} & 83.33\% & 10.00\% & 6.67\% \\
    \textbf{Content} & 13.33\% & 70.00\% & 16.67\% \\
    \textbf{Fluency} & 50.00\% & 50.00\% & 0.00\% \\
    \textbf{Coherence} & 40.00\% & 56.67\% & 3.33\% \\
    \textbf{Readability} & 86.67\% & 10.00\% & 3.33\% \\
    \bottomrule
  \end{tabular}
  \caption{\label{tab:single_model_results}
  Manual pair-wise comparison for 30 single document revisions without Meaning-changed edits.
  \vspace{-3mm}
  }
\end{table}
\begin{table}[t]
  \centering
  \small	
  \begin{tabular}{@{}c|ccc@{}}
    \toprule
    \textbf{$t$} & \textbf{Human Revisions} & \textbf{Tie} & \textbf{Model Revisions} \\
    \midrule
    1 & 57.14\% & 14.28\% & 28.58\% \\
    2 & 57.14\% & 14.28\% & 28.58\% \\
    3 & 42.85\% & 57.15\% & 0.00\% \\
    \bottomrule
  \end{tabular}
  \caption{\label{tab:iterative_model_results}
  Manual pair-wise comparison for overall quality of 21 iterative document-revisions, where $t$ indicates the revision depth.
  }
\end{table}

\paragraph{Manual Evaluation.}
Table \ref{tab:single_model_results} shows how the model revision affects the quality of the original document. 
We choose \textsc{Pegasus} trained on \textsc{IteraTeR-full} to generate revisions and compare with human revisions, as the model produces the best overall results\footnote{We provide detailed manual evaluation configuration in Appendix \ref{sec:human_eval_model_output}.}.
There exists a big gap between the best-performing model revisions and human revisions, indicating the challenging nature of the modeling problem.
Thus, while model revisions can achieve comparable performance with human revisions on fluency, coherence and meaning preservation, human revisions still outperform in terms of readability and overall quality.

Table \ref{tab:iterative_model_results} demonstrates how model-generated text quality varies across revision depths.
In the first two depths, human revisions win over model revisions with a ratio of 57.14\%.
However, in the last depth, \change[WD]{model revisions tie with human revisions with a similar percentage}{model revisions stay similar with human revisions in a ratio of 57.15\%}.
Upon reviewing revisions in the last depth, we find a lot of \change[WD]{content updates}{\textsc{Meaning-changed} edits} in human revisions. At the same time, the model revisions only made a few \textsc{Fluency} or \textsc{Clarity} edits, which the human evaluators tend to judge as ``tie''.

\begin{figure}[t]
    \includegraphics[width=\columnwidth]{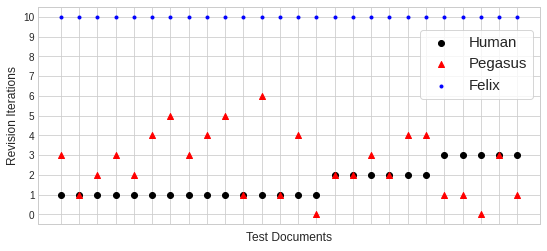}
    \caption{Number of iterations made by humans and different text revision models.}
    \label{fig:iterativeness}
\end{figure}

\paragraph{Iterativeness.}
We also compare the iterative ability between the two kinds of text revision models (best performing versions of both \textsc{Felix} and \textsc{Pegasus}: trained on \textsc{IteraTeR-full}), against human's iterative revisions. Figure \ref{fig:iterativeness} shows that while \textsc{Pegasus} is able to finish iterating after 2.57 revisions on average, \textsc{Felix} continues to make iterations until the maximum cutoff of 10 that we set for the experiment. In contrast, humans on average make 1.61 iterations per document. 
While \textsc{Felix} is able to make meaningful revisions (as evidenced by the improvements in the SARI metric in Table \ref{tab:model-performances}), it lacks the ability to effectively evaluate the text quality at a given revision, and decide whether or not to make further changes. \textsc{Pegasus}, on the other hand, is able to pick up on these nuances of iterative revision, and learns to stop revising after a certain level of quality has been reached.




\section{Conclusions and Discussions}

Our work is a step toward understanding the complex process of iterative text revision from human-written texts.
We collect, annotate and release \method: a novel, large-scale, domain-diverse, annotated dataset of human edit actions.
Our research shows that different domains of text have different distributions of edit intentions, and the general quality of the text has improved over time.
Computationally modeling the human's revision process is still under-explored, yet our results indicate some interesting findings and potential directions.

Despite the deliberate design of our dataset collection, \method only includes \textit{formally} written texts. We plan to extend it to diverse sets of revision texts, such as informally written blogs and less informal but communicative texts like emails, as well as increase the size of the current dataset.
For future research, we believe \method can serve as a basis for future corpus development and computationally modeling iterative text revision.

\section{Ethical Considerations}
We collect all data from publicly available sources, and respect copyrights for original document authors.
During the data annotation process, all human annotators are anonymized to respect their privacy rights.
We provide fair compensation to all human annotators, where each annotator gets paid more than the minimum wage and based on the number of annotations they conducted.


Our work has no possible harms to fall disproportionately on marginalized or vulnerable populations.
Our dataset does not contain any identity characteristics (e.g. gender, race, ethnicity), and will not have ethical implications of categorizing people.

\section*{Acknowledgments}
We thank all linguistic expert annotators at Grammarly for annotating, evaluating and providing feedback during our data annotation and evaluation process.
We appreciate that Courtney Napoles and Knar Hovakimyan at Grammarly helped coordinate the annotation resources.
We also thank Yangfeng Ji at University of Virginia and the anonymous reviewers for their helpful comments.

\bibliography{anthology,acl2021}
\bibliographystyle{acl_natbib}

\clearpage

\appendix
\section{Details on Text Processing in \method}
\label{appendix:data_processing}
For Wikipedia and Wikinews, we use the MediaWiki Action API\footnote{\url{https://www.mediawiki.org/wiki/API:Main_page}} to retrieve raw pages updated at different timestamps.
For each article, we start from July 2021 and trace back to its five most recent updated versions. 
Then, we parse\footnote{\url{https://github.com/earwig/mwparserfromhell}} plain texts from raw wiki-texts and filter out all references and external links.
For Wikipedia, we retrieve pages under the categories listed on the main category page
\footnote{\url{https://en.wikipedia.org/wiki/Wikipedia:Contents/Categories}}.
For Wikinews, we retrieve pages listed on the published articles page\footnote{\url{https://en.wikinews.org/wiki/Category:Published}}.

For ArXiv, we use the ArXiv API\footnote{\url{https://arxiv.org/help/api/}} to retrieve paper abstracts. 
Note that we do not retrieve the full paper for two reasons: (1) some paper reserved their copyright for distribution, (2) parsing and aligning editing actions in different document types (e.g. pdf, tex) is challenging. 
For each paper, we start from July 2021 and retrieve all its previous submissions.
We collect papers in the fields of Computer Science, Quantitative Biology, Quantitative Finance, and Economics.

\section{Details on Qualificiation Tests for Human Annotation}
\label{appendix:qualification_tests}
First, we prepare a small test set with 67 edit-actions and deploy parallel test runs on AMT to get more workers participate in this task.
Before starting the annotation, workers are required to pass a qualification test which has 5 test questions to get familiar with our edit-intention taxonomy.
Second, we compare workers' annotations with our golden annotations, and select workers who have an accuracy over 0.4.
After 5 test runs, we select 11 AMT workers who are qualified to participate in this task.
Then, we deploy the full 4K edit-actions on AMT, and collect 3 human annotations per edit-action.

\section{Human Annotation Instruction and Interface}
\label{sec:annotation-interface}
\begin{figure*}[h]
    \centering
    \includegraphics[width=\textwidth]{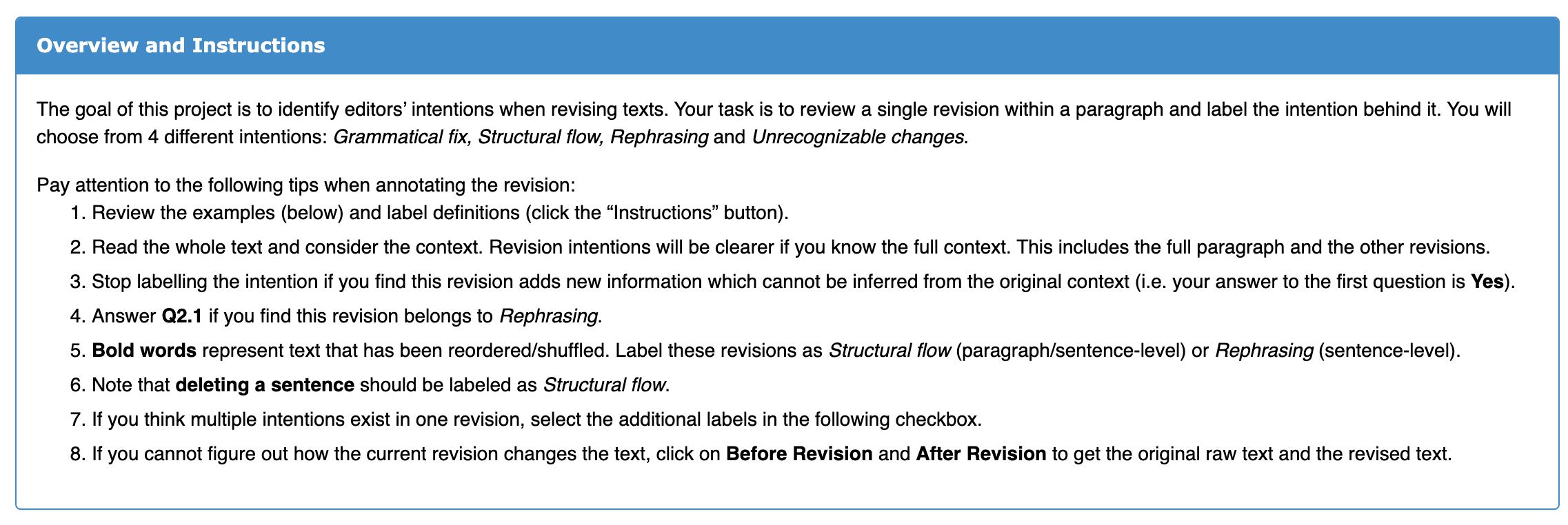}
    \caption{A screenshot of the annotation instruction for human annotators.}
    \label{fig:annotation_instrution}
\end{figure*}

\begin{figure*}[h]
    \centering
    \includegraphics[width=\textwidth]{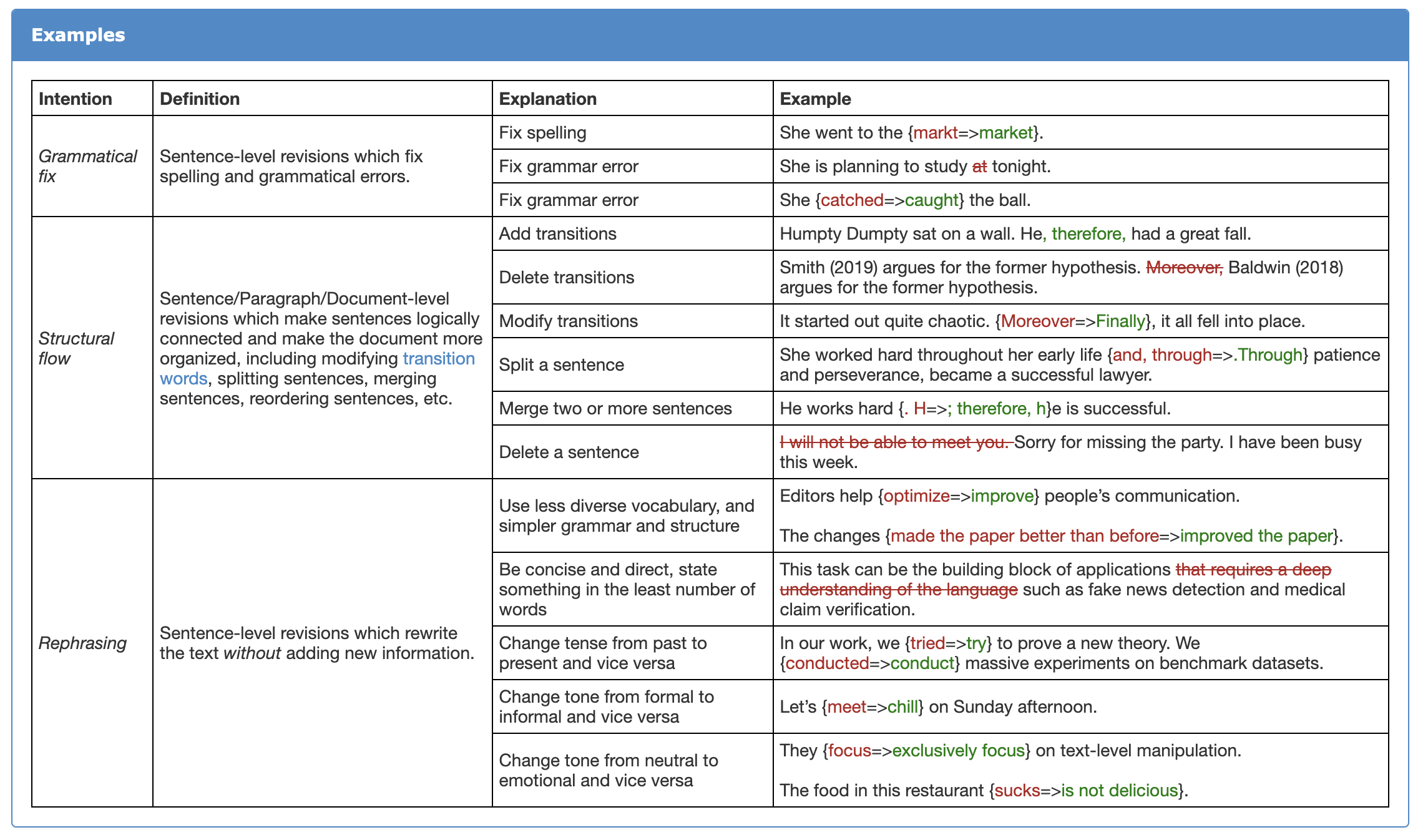}
    \caption{A screenshot of the provided examples for human annotators.}
    \label{fig:annotation_examples}
\end{figure*}

\begin{figure*}[h]
    \centering
    \includegraphics[width=\textwidth]{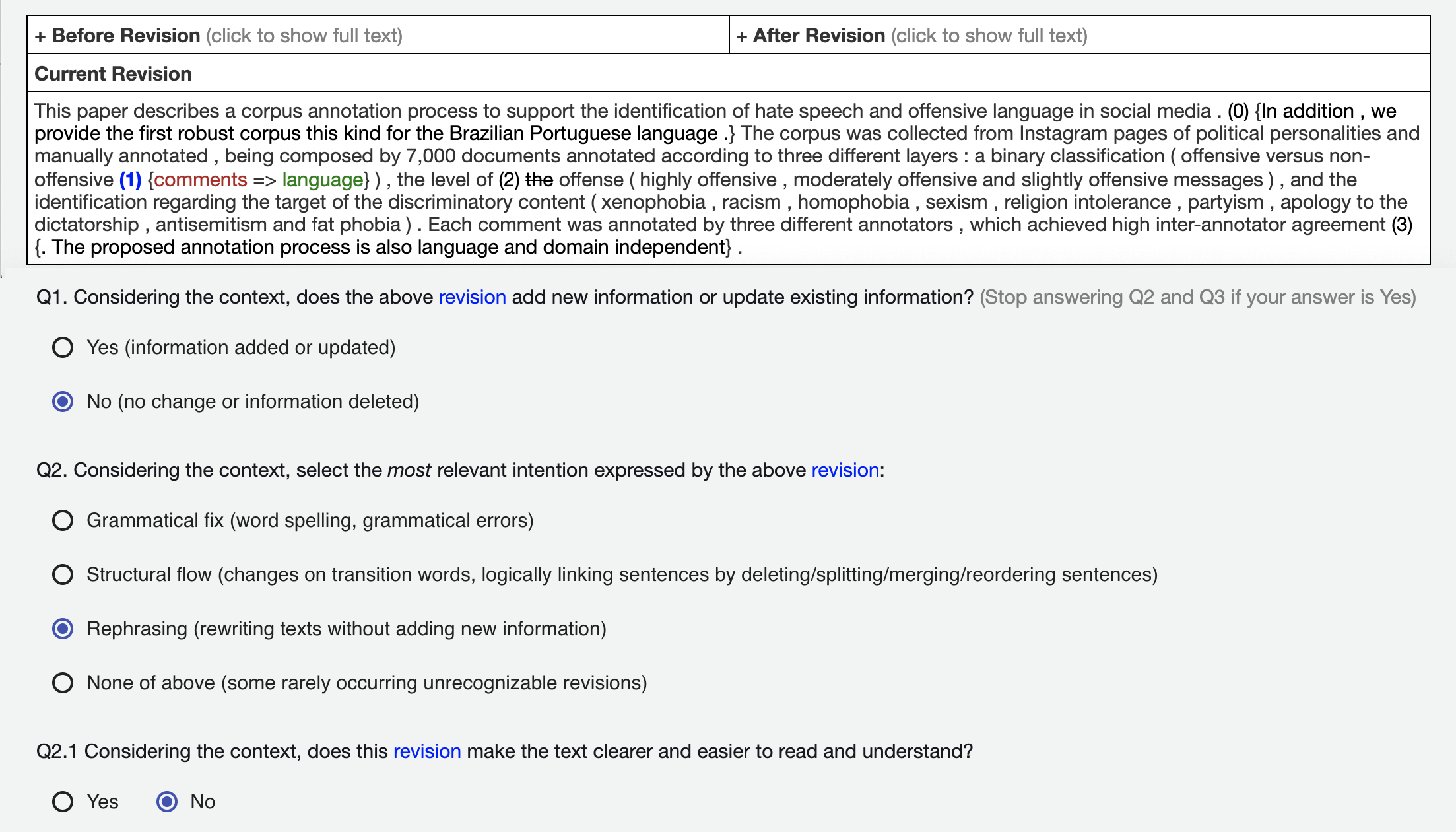}
    \caption{A screenshot of the annotation interface for human annotators.}
    \label{fig:annotation_questions}
\end{figure*}

To guide human annotators make accurate edit-intention annotation, we provide them with a short task instruction (Figure \ref{fig:annotation_instrution}) followed by some concrete edit-intention examples (Figure \ref{fig:annotation_examples}).
Then, we highlight the edit-action within the document-revision and ask human annotators three questions to obtain the accurate edit-intention of the current edit-action, as illustrated in Figure \ref{fig:annotation_questions}.
Note that in our previous test runs on AMT, we find that AMT workers can hardly have a consensus on Clarity and Style edits, which give a very low IAA score.
Therefore, in the annotation interface, we include Clarity and Style edits under the category of "Rephrasing", and further ask the annotators to judge whether the current "Rephrasing" edit is making the text more clearer and understandable. If yes, we convert this edit to Clarity, otherwise we convert this edit to Style.
This interface configuration gives us the best IAA score among our 5 test runs.

\section{Details on Computational Experiments}\label{appendix:auto_ann_details}
For all computational experiments in this work, we deploy them on a single Quadro RTX 4000(16GB) GPU.

\paragraph{RoBERTa.}
We leverage the RoBERTa-large model from Huggingface transformers \citep{wolf-etal-2020-transformers}, which has 354 million parameters.
We set the total training epoch to 15 and batch size to 4.
We use the Adam optimizer with weight decay \citep{loshchilov2018decoupled}, and set the learning rate to $10^{-5}$ which decreases linearly to 0 at the last training iteration.
We report descriptive statistics with a single run.
We use the sklearn package \citep{scikit-learn} to calculate the precision, recall and f1 score.

\paragraph{Text Revision Models.}
We leverage the BART-large (with 400 million parameters) and PEGASUS-large (with 568 million parameters) from Huggingface transformers \citep{wolf-etal-2020-transformers}.
We set the total training epoch to 5 and batch size to 16.
We use the Adam optimizer with weight decay \citep{loshchilov2018decoupled}, and set the learning rate to $3\times10^{-5}$ which decreases linearly to 0 at the last training iteration.
We report descriptive statistics with a single run.
We use the metrics package from Huggingface transformers to calculate the SARI, BLEU, ROUGE-1/2/L score.

\section{Justification of Automatic Evaluation Metrics}\label{appendix:auto_eval_metrics_justification}
For \textbf{Fluency}, we use the Syntactic Log-Odds Ratio (SLOR) \citep{kann-etal-2018-sentence} to evaluate the naturalness and grammaticality of the current revised document, where a higher SLOR score indicates a more fluent document. 
Prior works \citep{pauls-klein-2012-large,kann-etal-2018-sentence} found word-piece log-probability correlates well with human fluency ratings. 
For \textbf{Coherence}, we use the Entity Grid (EG) score \citep{lapata2005automatic} to evaluate the local coherence of the current revised document, where a higher EG score indicates a more coherent document. EG is a widely adopted \citep{soricut-marcu-2006-discourse,elsner-charniak-2008-coreference,louis-nenkova-2012-coherence} metric for measuring document coherence.
For \textbf{Readability}, we use the the Flesch–Kincaid Grade Level (FKGL) \citep{kincaid1975derivation} to evaluate how easy the current revised document is for the readers to understand, where a lower FKGL indicates a more readable document. FKGL is a popular metric that has been used by many prior works \citep{solnyshkina2017evaluating,10.1162/tacl_a_00107,guo-etal-2018-dynamic,nassar-etal-2019-neural,nishihara-etal-2019-controllable} to measure the readability of documents.
For \textbf{Content Preservation}, we use the BLEURT score \citep{sellam-etal-2020-bleurt} to measure how much content has been changed from the previous document to the current revised one, where a higher BLEURT score indicates more content has been preserved. BLEURT has been shown to correlate better with human judgments than other metrics that take semantic information into account, e.g. METEOR \citep{banerjee-lavie-2005-meteor} or BERTScore \citep{bert-score}. 

\section{Details on Human Evaluation for Single Human Revision Quality}
\label{sec:human_eval_single_doc}
\begin{table}[t]
  \centering
  \scriptsize
  \begin{tabular}{@{}c|cccc@{}}
    \toprule
    \textbf{Overall} $\uparrow$ & \textbf{BLEURT}$\uparrow$ & \textbf{$\Delta$SLOR} $\uparrow$ & \textbf{$\Delta$EG} $\uparrow$ & \textbf{$\Delta$FKGL} $\downarrow$\\
    \midrule
    0.5800 & 0.4709 & -0.0757 & -0.0098 & -0.6301 \\
    \bottomrule
  \end{tabular}
  \caption{\label{tab:single_revision_metrics}
  Evaluation results for 50 document-revisions.
  Note that $\Delta$SLOR, $\Delta$EG and $\Delta$FKGL are computed by subtracting the scores of original documents from the scores of revised documents.
  \vspace{-3mm}
  }
\end{table}

\paragraph{Evaluation Data.}
To evaluate how do human revisions affect the text quality, we sample 50 single document-revisions, which contains 50 randomly sampled documents and each document has 1 document-revision.\footnote{We exclude documents including Meaning-changed edits} 

\paragraph{Result Analysis.}
In Table \ref{tab:single_revision_metrics}, 
we observe that human revised documents generally improve the overall quality of original documents. 
As for the automatic metrics, BLEURT indicates that human revisions preserve much of the content, and $\Delta$FKGL shows that the readability of original documents improves by human revisions. However, $\Delta$SLOR and $\Delta$EG show a slight drop in performance.
We conjecture this is because (1) $\Delta$SLOR and $\Delta$EG are not well-aligned with human judgements, or (2) human revisions make original documents less fluent and less coherent.

\begin{table}[t]
  \centering
  \small
  \begin{tabular}{@{}l|r|r@{}}
    \toprule
    \multicolumn{1}{l|}{} & \multicolumn{2}{c}{\textbf{Human Overall}} \\
    \cmidrule(lr){2-3}
    \multicolumn{1}{l|}{} & \multicolumn{1}{c|}{\textbf{Pearson}} & \multicolumn{1}{c}{\textbf{Spearman}}\\
    \midrule
    \textbf{BLEURT} & 0.1139 (0.3626) & 0.0756 (0.5465) \\
    \textbf{$\Delta$SLOR} & -0.1239 (0.3216) & -0.2218 (0.0734) \\
    \textbf{$\Delta$EG} & -0.1480 (0.2355) & 0.0187 (0.8817) \\
    \textbf{$\Delta$FKGL} & 0.1171 (0.3491) & 0.2042 (0.1001) \\
    \bottomrule
  \end{tabular}
  \caption{\label{tab:metric_human_corr}
  Correlation coefficients between human overall score and automatic metrics, where numbers in the parentheses is the $p$-value.
  }
\end{table}

\paragraph{Correlation Analysis.}
To analyze how automatic metrics are correlated with human overall quality score, we compute the Pearson \citep{kowalski1972effects} and Spearman \citep{zwillinger1999crc} correlation coefficients between the automatic metrics and the human overall quality scores based on 50 single document-revisions and 21 iterative document-revisions.
Table \ref{tab:metric_human_corr} shows that BLEURT and $\Delta$FKGL are positively correlated with human overall quality score, while $\Delta$SLOR and $\Delta$EG are negatively correlated with human overall quality score.

\section{Details on Human Evaluation Configuration for Model Revisions}
\label{sec:human_eval_model_output}

\begin{figure*}[h]
    \centering
    \includegraphics[width=\textwidth]{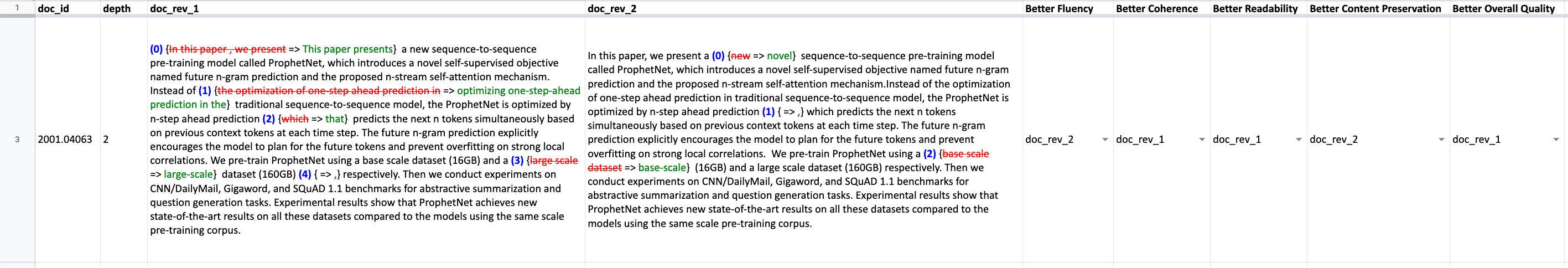}
    \caption{A screenshot of the single document-revision quality evaluation interface for human evaluators.}
    \label{fig:eval_single_interface}
\end{figure*}

First, we evaluate how do model revisions affect the quality of the document.
We randomly sample 30 single document-revisions which do not contain Meaning-changed edits, and input the original documents to the best-performing model to get the model-revised documents.
Then, for each data pair, we randomly shuffle model revisions and human revisions, and ask human evaluators to select which revision leads to better document quality in terms of:
\begin{itemize}[noitemsep,topsep=0pt,leftmargin=*]
\item \textbf{Content Preservation}: keeping more content information unchanged;
\item \textbf{Fluency}: fixing more grammatical errors or syntactic errors;
\item \textbf{Coherence}: making the sentences more logically linked and organized;
\item \textbf{Readability}: making the text easier to read and understand;
\item \textbf{Overall Quality}: better improving the overall quality of the document.
\end{itemize}
We provide the evaluation interface in Figure \ref{fig:eval_single_interface}.

\begin{figure*}[h]
    \centering
    \includegraphics[width=\textwidth]{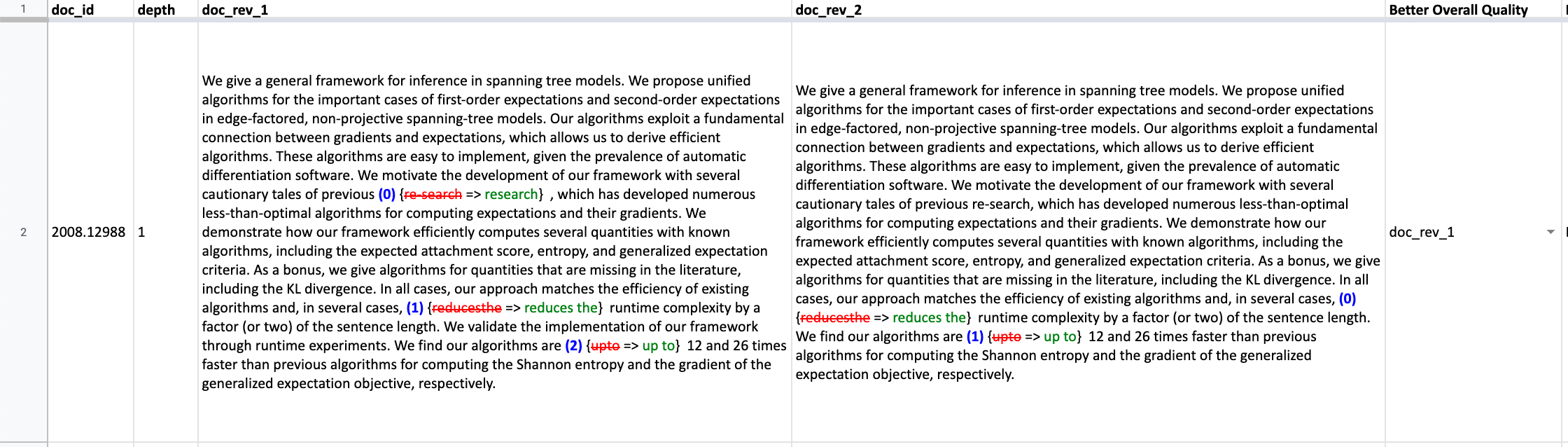}
    \caption{A screenshot of the iterative document-revision quality evaluation interface for human evaluators.}
    \label{fig:eval_iterative_interface}
\end{figure*}

Secondly, we evaluate how does model generated text quality vary across revision depths.
We use the same set of 21 iterative document-revisions in \S\ref{sec:human-data-setup}.
We feed the original documents into the best-performing model to obtain the model revised documents at each revision depth.
For each data pair, we randomly shuffle model revisions and human revisions, and ask human evaluators to judge which one gives better overall text quality.
We provide the evaluation interface in Figure \ref{fig:eval_iterative_interface}.

\begin{table*}[t]
  \centering
  \small
  \begin{tabular}{@{}ll|cccc|cccc|c@{}}
    \toprule
    \textbf{Model} & \textbf{Training Data} & \textbf{SARI} & \textbf{DEL} & \textbf{ADD} & \textbf{KEEP} & 
    \textbf{BLEU} & \textbf{R-1} & \textbf{R-2} & \textbf{R-L} & \textbf{Avg.}\\
    \midrule
    \textsc{Felix} & \textsc{IteraTeR-human-raw} & 29.23 & 19.23 & 0.62 & 67.85 & 49.48 & 77.27 & 60.11 & 63.43 & 47.38\\
    \textsc{Felix} & \textsc{IteraTeR-human} & 30.65 & 20.26 & 0.99 & 70.71 & 54.35 & 78.97 & 58.46 & 59.06 & 48.02\\
    \textsc{Felix} & \textsc{IteraTeR-full-raw} & 30.34 & 20.44 & 1.40 & 69.18 & 55.10 & 76.47 & 58.07 & 56.49 & 47.31\\
    \textsc{Felix} & \textsc{IteraTeR-full} & \textbf{33.48} & \textbf{22.39} & \textbf{2.52} & \textbf{75.52} & \textbf{61.90} & \textbf{80.65} & \textbf{64.97} & \textbf{63.72} & \textbf{53.03} \\
    \midrule
    \midrule    
    \textsc{BART} & \textsc{IteraTeR-human-raw} & 33.20 & 9.81 & 3.58 & 86.20 & \textbf{78.59} & 85.93 & 79.94 & 85.20 & 65.66\\
    \textsc{BART} & \textsc{IteraTeR-human} & 34.77 & 13.43 & \textbf{5.91} & 84.97 & 74.43 & 85.23 & 79.00 & 84.45 & 64.55\\
    \textsc{BART} & \textsc{IteraTeR-full-raw} & 33.88 & 12.38 & 2.34 & \textbf{86.92} & 78.55 & 86.66 & \textbf{80.97} & 86.05 & 66.16 \\
    \textsc{BART} & \textsc{IteraTeR-full} & \textbf{37.28} & \textbf{19.83} & 5.69 & 86.33 & 77.50 & \textbf{86.85} & 80.43 & \textbf{86.14} & \textbf{66.97} \\
    \midrule
    \textsc{Pegasus} & \textsc{IteraTeR-human-raw} & 33.09 & 10.61 & 1.57 & 87.09 & 
        \textbf{79.09} & 87.50 & 81.65 & 86.77 & 66.32 \\
    \textsc{Pegasus} & \textsc{IteraTeR-human} & 34.43 & 13.26 & 2.89 & 87.14 & 
        78.85 & 87.53 & 81.77 & 86.84 & 66.71\\
    \textsc{Pegasus} & \textsc{IteraTeR-full-raw} & 34.67 & 13.93 & 2.36 & \textbf{87.53} & 
        78.21 & \textbf{87.63} & \textbf{82.02} & \textbf{87.06} & 66.65 \\
    \textsc{Pegasus} & \textsc{IteraTeR-full} & \textbf{37.11} & \textbf{19.66} & \textbf{4.44} & 87.16 &
        77.60 & 87.42 & 81.84 & 86.84 & \textbf{67.18} \\
    \midrule
    Baseline & - & 29.47 & 0.0 & 0.0 & 88.42 & 81.25 & 88.67 & 83.51 & 88.04 & 66.25 \\
    \bottomrule
  \end{tabular}
  \caption{\label{tab:model-performances}
  Model performances evaluated on the test set of \textsc{IteraTeR-human}. \change[ZM]{R-1, R-2 and R-L refer to ROUGE-1, ROUGE-2, and ROUGE-L metrics, respectively.}{R-1, R-2, and R-L refer to ROUGE-1, ROUGE-2, and ROUGE-L metrics, respectively, and Avg is computed by taking the mean of SARI, BLEU, and R-L scores.} Baseline refers to a no-edit baseline, where we simply use the input text as the output. 
  }
\end{table*}
\begin{table}[t]
  \centering
  \small
  \begin{tabular}{@{}cl|cccc@{}}
    \toprule
    $t$ & \textbf{Edit-Intention} & \textbf{SARI} & \textbf{ADD} & \textbf{DEL} & \textbf{KEEP} \\
    \midrule
    1 & \shading{Fluency} & 46.22 & 18.53 & 24.00 & 96.12  \\
    1 & \shading{Coherence} & 38.33 & 6.42 & 17.91 & 90.66 \\
    1 & \shading{Clarity} & 34.35 & 1.72 & 1.72 & 82.54 \\
    1 & \shading{Style} & 40.61 & 0.0 & 32.63 & 89.19 \\
    \midrule
    2 & \shading{Fluency} & 30.71 & 0.0 & 0.0 & 92.14 \\
    2 & \shading{Coherence} & 29.50 & 0.0 & 6.25 & 82.26 \\
    2 & \shading{Clarity} & 35.29 & 7.33 & 18.94 & 86.19 \\
    2 & \shading{Style} & 30.34 & 0.0 & 0.0 & 91.04 \\
    \midrule
    3 & \shading{Fluency} & 32.74 & 0.0 & 3.98 & 94.26 \\
    3 & \shading{Coherence} & 37.18 & 0.0 & 21.13 & 90.41 \\
    3 & \shading{Clarity} & 34.62 & 0.0 & 21.04 & 82.81 \\
    3 & \shading{Style} & 32.09 & 0.0 & 37.50 & 58.77 \\
    \bottomrule
  \end{tabular}
  \caption{\label{tab:sari-details}
  Detailed SARI scores for \textsc{Pegasus} trained on \textsc{IteraTeR-full} and evaluated on the test set of \textsc{IteraTeR-human}, where $t$ is the revision depth.
  }
\end{table}

\section{Details on Automatic Evaluation for Model Revisions}
\label{sec:auto_eval_details}
Table \ref{tab:model-performances} provides detailed automatic evaluation results for \textsc{Felix} and \textsc{Pegasus}, including SARI, BLEU, and ROUGE.
We choose these automatic metrics following prior text revision works \citep{malmi-etal-2019-encode,dong-etal-2019-editnts,mallinson-etal-2020-felix}.
\add[WD]{Note that the KEEP score of Baseline is not 100 because the source sentence keeps all n-grams, but there might be certain n-grams that are not kept in the reference sentence. This results in the non-perfect KEEP score since both recall and precision are calculated.}

Table \ref{tab:sari-details} further provides SARI score under different revision depths as well as different edit-intentions.
We find that \textsc{Pegasus} only conduct deletions in the revision depth 3, and the SARI score for each edit-intention varies a lot across different revision depths.

Table \ref{tab:model-output-example} and Table \ref{tab:model-output-example2} are some examples of iterative text revisions generated by \textsc{Felix} and \textsc{Pegasus} trained on \textsc{IteraTeR-full}.
We observe that while \textsc{Felix} can make more edits with more iterations than \textsc{Pegasus}, it cannot ensure the quality of its generated edits. 
\textsc{Felix} often insert some random out-of-context tokens into the original text, and distort the semantic meaning of the original text.
\textsc{Pegasus} is better at preserving the semantic meaning of the original text, but it is more likely to delete phrases or tokens in deeper revision depth.

\begin{table*}[t]
  \centering
  \small
  \begin{tabular}{@{}r|p{0.3\textwidth}|p{0.3\textwidth}|p{0.3\textwidth}@{}}
    \toprule
     $t$ & \textsc{Felix} & \textsc{Pegasus} & \textsc{Human} \\
    \midrule
    0 & The three shareholders had unanimously agreed not to appoint an auditor for the company , but according to Investigate Magazine, another shareholder  Russell Hyslop, had never been consulted about the matter. & The three shareholders had unanimously agreed not to appoint an auditor for the company , but according to Investigate Magazine, another shareholder  Russell Hyslop, had never been consulted about the matter. & The three shareholders had unanimously agreed not to appoint an auditor for the company , but according to Investigate Magazine, another shareholder  Russell Hyslop, had never been consulted about the matter.\\
    \midrule
    1 & \textcolor{red}{\sout{T}}\textcolor{teal}{t}he three shareholders had unanimously agreed not to appoint an auditor for the company , but according to \textcolor{red}{\sout{I}}\textcolor{teal}{i}nvestigate \textcolor{red}{\sout{M}}\textcolor{teal}{m}agazine , \textcolor{red}{\sout{another shareholder  Russell Hyslop}}, had never been consulted about the matter\textcolor{teal}{ , another }. & The three shareholders had unanimously agreed not to appoint an auditor for the company\textcolor{red}{\sout{,}}\textcolor{teal}{.}   \textcolor{red}{\sout{but a}}\textcolor{teal}{A}ccording to Investigate Magazine, another shareholder Russell Hyslop, had never been consulted about the matter. & The three shareholders had unanimously agreed not to appoint an auditor for the company\textcolor{red}{\sout{,}}\textcolor{teal}{;} but according to Investigate Magazine, another shareholder Russell Hyslop, had never been consulted about the matter. \\
    \midrule
    2 & the three shareholders had unanimously agreed not to appoint an auditor for the company , but according to \textcolor{red}{\sout{investigate magazine}}, had never been consulted about the matter , another \textcolor{teal}{investigate} . & The three shareholders had unanimously agreed not to appoint an auditor for the company. According to \textcolor{red}{\sout{Investigate Magazine}}, another shareholder Russell Hyslop, had never been consulted about the matter. & The three shareholders had unanimously agreed not to appoint an auditor for the company\textcolor{red}{\sout{;}}\textcolor{teal}{,} but according to Investigate Magazine, another shareholder\textcolor{teal}{,} Russell Hyslop, had never been consulted about the matter. \\
    \midrule
    3 & the three shareholders had unanimously agreed not to appoint an auditor for the company , \textcolor{teal}{had never been consulted about the matter ,} but according to \textcolor{teal}{reuters , no . g launched} \textcolor{red}{\sout{had never been consulted about the matter ,}} another investigate . & The three shareholders had unanimously agreed not to appoint an \textcolor{teal}{external} auditor for the company. According to, another shareholder Russell Hyslop, had never been consulted about the matter. & - \\
    \midrule
    4 & the three shareholders had unanimously agreed not to appoint an auditor for the company , had never been consulted about the matter , but according to \textcolor{red}{\sout{reuters}}, no . g launched another investigate . & The three shareholders had unanimously agreed not to appoint an external auditor for the company. According to\textcolor{red}{\sout{,}} another shareholder Russell Hyslop, had never been consulted about the matter. & - \\
    \midrule
    5 & the three shareholders had unanimously agreed not to appoint an auditor for the company , had never been consulted about the matter , but according to \textcolor{red}{\sout{, no . g }}\textcolor{teal}{cre et company . the company} launched another investigate \textcolor{red}{\sout{.}}\textcolor{teal}{inc . , no . g} & - & -\\
    \midrule
    6 & the three shareholders had unanimously agreed not to appoint an auditor for the company , had never been consulted about the matter , but \textcolor{red}{\sout{according to cre et company . the company}} launched another investigate inc . , no . g . & - & -\\
    \bottomrule
  \end{tabular}
  \caption{\label{tab:model-output-example}
  A sample snippet of iterative text revisions generated by \textsc{Felix} and \textsc{Pegasus} (both trained on \textsc{IteraTeR-full}, conditioned on Fluency intention) respectively, where $t=0$ indicates the original input text.
  \vspace{-3mm}
  }
\end{table*}

\begin{table*}[t]
  \centering
  \small
  \begin{tabular}{@{}r|p{0.3\textwidth}|p{0.3\textwidth}|p{0.3\textwidth}@{}}
    \toprule
     $t$ & \textsc{Felix} & \textsc{Pegasus} & \textsc{Human} \\
    \midrule
    0 & Trainee programs and graduate programs are arranged by private companies and public sector employers where the trainee is offered the possibility to take part 6 to 20 months training programs . During the duration of these programs, the trainee is expected to receive a salary as well as is expected to have full-time employment awaiting in the company when the program is over. & Trainee programs and graduate programs are arranged by private companies and public sector employers where the trainee is offered the possibility to take part 6 to 20 months training programs . During the duration of these programs, the trainee is expected to receive a salary as well as is expected to have full-time employment awaiting in the company when the program is over. & Trainee programs and graduate programs are arranged by private companies and public sector employers where the trainee is offered the possibility to take part 6 to 20 months training programs . During the duration of these programs, the trainee is expected to receive a salary as well as is expected to have full-time employment awaiting in the company when the program is over.\\
    \midrule
    1 & \textcolor{red}{\sout{T}}\textcolor{teal}{jobs , t}raining and graduate programs are arranged by private companies and public sector employ\textcolor{red}{\sout{ers}} \textcolor{red}{\sout{where the trainee is offered the possibility to take part 6 to 20 months training programs . During the duration of these programs, the trainee is expected to receive a salary as well as is expected to have full-time employment awaiting in the company}}\textcolor{teal}{ment awaiting in full - time salary as well as is expected to receive a duration of these programs , the company} when the program is over . \textcolor{teal}{employers where the trainee programs is expected to have part time . during training the trainee as the trainees assigned as the} 
    & Trainee programs and graduate programs are arranged by private companies and public sector employers where the trainee is offered the possibility to take part \textcolor{red}{\sout{6}}\textcolor{teal}{six} to 20 months training programs. During the duration of these programs, the trainee is expected to receive a salary as well as is expected to have full-time employment awaiting in the company when the program is over. 
    & Trainee programs \textcolor{red}{\sout{and graduate programs}} are arranged by private companies and public sector employers where the trainee is offered the possibility to take part 6 to 20 months training programs . During the duration of these programs, the trainee is expected to receive a salary as well as is expected to have full-time employment awaiting in the company when the program is over. \\
    \midrule
    2 & jobs, training and graduate programs are arranged by private companies and public sector employment awaiting in full - time salary as well as is expected to receive a duration of these programs , the company when \textcolor{red}{\sout{the program}} is \textcolor{red}{\sout{over}}. \textcolor{red}{\sout{ employ-ers  where  the  trainee  programs  is  expected  tohave part time . during training the trainee as thetrainees assigned}}. as the 
    & Trainee programs and graduate programs are arranged by private companies and public sector employers\textcolor{teal}{,} where the trainee is offered the possibility to take part six to 20 months training programs. During the duration of these programs, the trainee is expected to receive a salary as well as is expected to have full-time employment awaiting in the company when the program is over. 
    & Trainee programs \textcolor{teal}{and graduate programs} are arranged by private companies and public sector employers where the trainee \textcolor{red}{\sout{is offered the possibility to take part 6 to 20 months training programs}} \textcolor{teal}{position has a varied duration depending on the company's program} . During the duration of these programs, the trainee is expected to receive a salary as well as is expected to have full-time employment awaiting in the company when the program is over. \\
    \midrule
    3 & jobs , training and graduate programs are arranged by private companies and public sector employment awaiting \textcolor{red}{\sout{ in  full - time salary as well as is expected to receive}} a duration of these programs \textcolor{red}{\sout{the  company  when . as the}}\textcolor{teal}{, full - time salary expected to receive the company when received as well} 
    & Training programs and graduate programs are organized by private companies and public sector employers, where the trainee \textcolor{red}{\sout{is offered the possibility to take part six to 20 months training programs. During the duration of these programs, the trainee}} is expected to receive a salary as well as is expected to have full-time employment awaiting in the company when the program is over. & - \\
    \midrule
    4 & jobs , training and graduate programs are arranged by private companies and public sector employment awaiting a duration of these programs , full - time salary expected to receive the company when received \textcolor{red}{\sout{as well}} 
    & Training programs and graduate programs are organized by private companies and public sector employers, where the trainee is expected to receive a salary and \textcolor{red}{\sout{as well as}} is expected to have full-time employment awaiting in the company when the program is over. & - \\
    \bottomrule
  \end{tabular}
  \caption{\label{tab:model-output-example2}
  A sample snippet of iterative text revisions generated by \textsc{Felix} and \textsc{Pegasus} (both trained on \textsc{IteraTeR-full}, conditioned on Clarity intention) respectively, where $t=0$ indicates the original input text.
  \vspace{-3mm}
  }
\end{table*}

\end{document}